\documentclass[11pt,a4paper]{article}
\usepackage[hyperref]{acl2019}
\usepackage{times}
\usepackage{latexsym}
\usepackage{url}
\usepackage{graphicx}
\usepackage{amsmath}
\usepackage{amssymb}
\usepackage{subfigure}
\usepackage{caption}
\usepackage{algorithm}
\usepackage{algorithmicx}
\usepackage{algpseudocode}
\usepackage{multicol}
\usepackage{multirow}
\usepackage{booktabs}
\usepackage{listings}
\usepackage{array}
\usepackage{bm}

\aclfinalcopy 

\title{ERNIE: Enhanced Language Representation with Informative Entities}

 \author{Zhengyan Zhang$^{1,2,3*}$, Xu Han$^{1,2,3*}$, Zhiyuan Liu$^{1,2,3\dagger}$, Xin Jiang$^4$, Maosong Sun$^{1,2,3}$, Qun Liu$^4$ \\
        $^1$Department of Computer Science and Technology, Tsinghua University, Beijing, China  \\
         $^2$Institute for Artificial Intelligence, Tsinghua University, Beijing, China \\ 
         $^3$State Key Lab on Intelligent Technology and Systems, Tsinghua University, Beijing, China\\
         $^4$Huawei Noah's Ark Lab \\
         \texttt{\{zhangzhengyan14,hanxu17\}@mails.tsinghua.edu.cn}\\
}

\begin{document}
\maketitle
\begin{abstract}

Neural language representation models such as BERT pre-trained on large-scale corpora can well capture rich semantic patterns from plain text, and be fine-tuned to consistently improve the performance of various NLP tasks. However, the existing pre-trained language models rarely consider incorporating knowledge graphs (KGs), which can provide rich structured knowledge facts for better language understanding. We argue that informative entities in KGs can enhance language representation with external knowledge. In this paper, we utilize both large-scale textual corpora and KGs to train an enhanced language representation model (ERNIE), which can take full advantage of lexical, syntactic, and knowledge information simultaneously. The experimental results have demonstrated that ERNIE achieves significant improvements on various knowledge-driven tasks, and meanwhile is comparable with the state-of-the-art model BERT on other common NLP tasks. The source code and experiment details of this paper can be obtained from \url{https://github.com/thunlp/ERNIE}.

\end{abstract}

\section{Introduction}

{\let\thefootnote\relax\footnotetext{$^*$ indicates equal contribution}}
{\let\thefootnote\relax\footnotetext{$^\dagger$ Corresponding author: Z.Liu(liuzy@tsinghua.edu.cn)}}

Pre-trained language representation models, including feature-based~\cite{mikolov2013distributed,pennington2014glove,peters2017semi,peters2018deep} and fine-tuning~\cite{dai2015semi,howard2018universal,radford2018improving,devlin2018bert} approaches, can capture rich language information from text and then benefit many NLP applications. BERT~\cite{devlin2018bert}, as one of the most recently proposed models, obtains the state-of-the-art results on various NLP applications by simple fine-tuning, including named entity recognition~\cite{sang2003introduction}, question answering~\cite{rajpurkar2016squad,zellers2018swag}, natural language inference~\cite{bowman2015large}, and text classification~\cite{wang2018glue}. 

\begin{figure}[t]
\centering
\includegraphics[width = 1.0\linewidth]{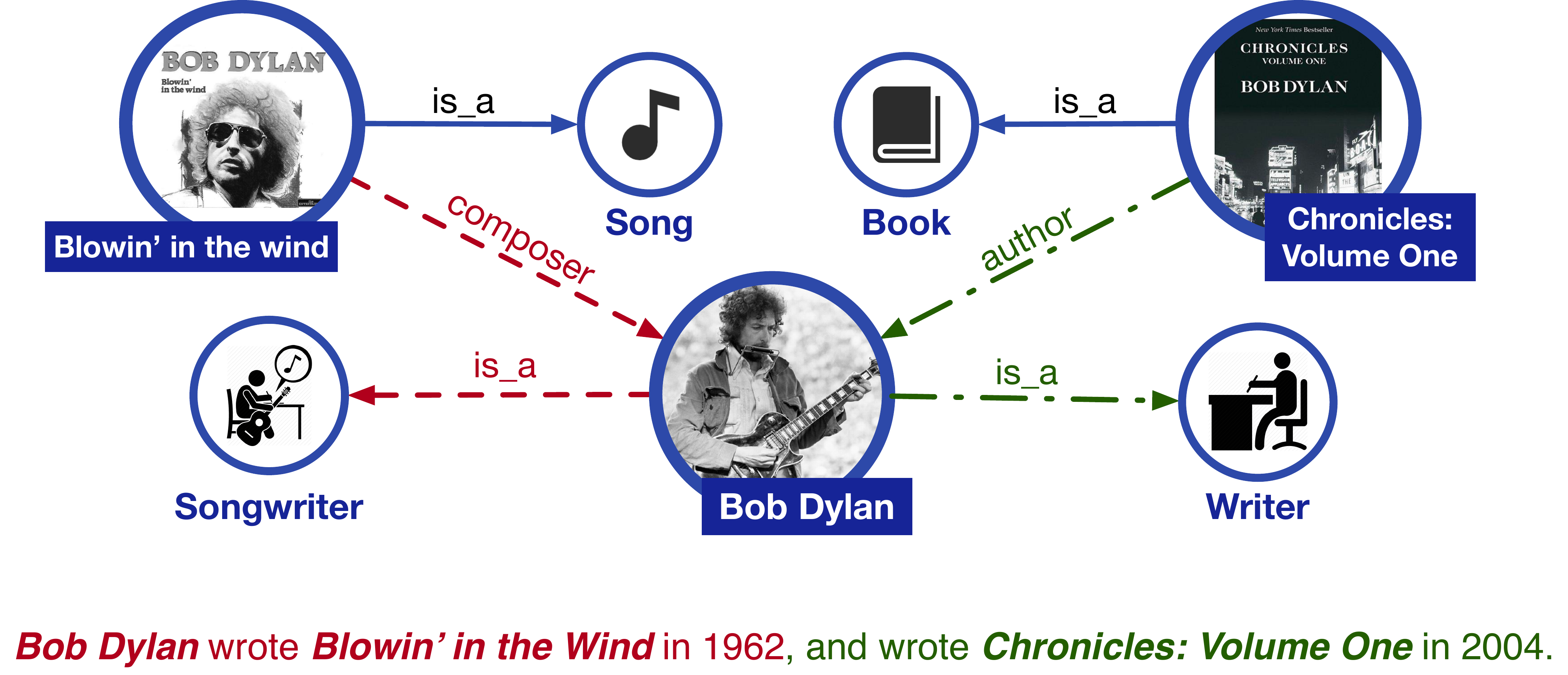}
\captionsetup{font={normalsize}}
\caption{An example of incorporating extra knowledge information for language understanding. The solid lines present the existing knowledge facts. The red dotted lines present the facts extracted from the sentence in red. The green dot-dash lines present the facts extracted from the sentence in green.} 
\label{fig:bob}
\vspace{-5.1mm}
\end{figure}

Although pre-trained language representation models have achieved promising results and worked as a routine component in many NLP tasks, they neglect to incorporate knowledge information for language understanding. As shown in Figure~\ref{fig:bob}, without knowing \emph{Blowin' in the Wind} and \emph{Chronicles: Volume One} are \emph{song} and \emph{book} respectively, it is difficult to recognize the two occupations of \emph{Bob Dylan}, i.e., \texttt{songwriter} and \texttt{writer}, on the entity typing task. Furthermore, it is nearly impossible to extract the fine-grained relations, such as \texttt{composer} and \texttt{author} on the relation classification task. For the existing pre-trained language representation models, these two sentences are syntactically ambiguous, like ``\emph{UNK} wrote \emph{UNK} in \emph{UNK}''. Hence, considering rich knowledge information can lead to better language understanding and accordingly benefits various knowledge-driven applications, e.g. entity typing and relation classification.

For incorporating external knowledge into language representation models, there are two main challenges. (1) \textbf{Structured Knowledge Encoding}: regarding to the given text, how to effectively extract and encode its related informative facts in KGs for language representation models is an important problem; (2) \textbf{Heterogeneous Information Fusion}: the pre-training procedure for language representation is quite different from the knowledge representation procedure, leading to two individual vector spaces. How to design a special pre-training objective to fuse lexical, syntactic, and knowledge information is another challenge.

To overcome the challenges mentioned above, we propose \textbf{E}nhanced Language \textbf{R}epresentatio\textbf{N} with \textbf{I}nformative \textbf{E}ntities (ERNIE), which pre-trains a language representation model on both large-scale textual corpora and KGs: 

(1) For extracting and encoding knowledge information, we firstly recognize named entity mentions in text and then align these mentions to their corresponding entities in KGs. Instead of directly using the graph-based facts in KGs, we encode the graph structure of KGs with knowledge embedding algorithms like TransE~\cite{bordes2013translating}, and then take the informative entity embeddings as input for ERNIE. Based on the alignments between text and KGs, ERNIE integrates entity representations in the knowledge module into the underlying layers of the semantic module.


(2) Similar to BERT, we adopt the masked language model and the next sentence prediction as the pre-training objectives. Besides, for the better fusion of textual and knowledge features, we design a new pre-training objective by randomly masking some of the named entity alignments in the input text and asking the model to select appropriate entities from KGs to complete the alignments. Unlike the existing pre-trained language representation models only utilizing local context to predict tokens, our objectives require models to aggregate both context and knowledge facts for predicting both tokens and entities, and lead to a knowledgeable language representation model.


We conduct experiments on two knowledge-driven NLP tasks, i.e., entity typing and relation classification. The experimental results show that ERNIE significantly outperforms the state-of-the-art model BERT on these knowledge-driven tasks, by taking full advantage of lexical, syntactic, and knowledge information. We also evaluate ERNIE on other common NLP tasks, and ERNIE still achieves comparable results.

\begin{figure*}[t]
\centering
\includegraphics[width = \linewidth]{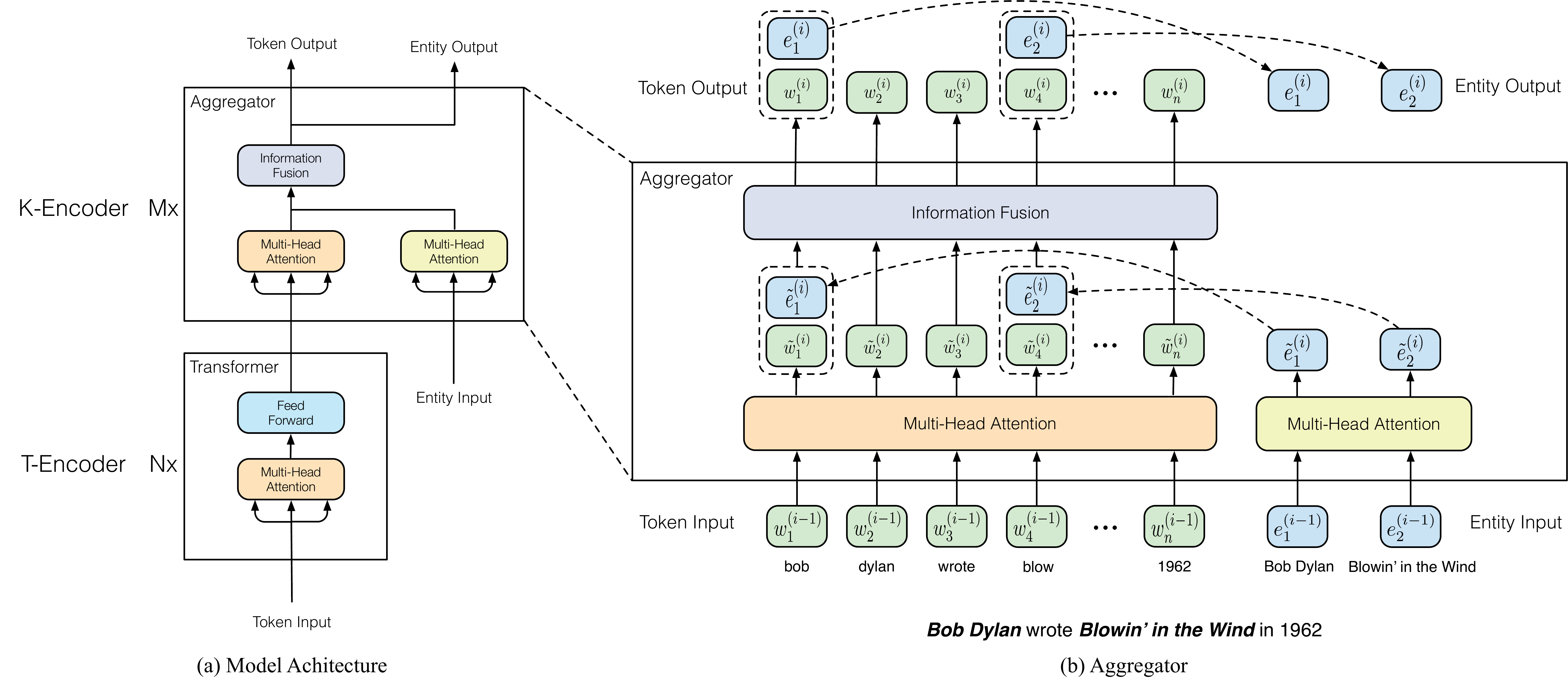}
\captionsetup{font={normalsize}}
\caption{The left part is the architecture of ERNIE. The right part is the aggregator for the mutual integration of the input of tokens and entities. 
Information fusion layer takes two kinds of input: one is the token embedding, and the other one is the concatenation of the token embedding and entity embedding. After information fusion, it outputs new token embeddings and entity embeddings for the next layer.}
\label{fig:fusion}
\end{figure*}

\section{Related Work}

Many efforts are devoted to pre-training language representation models for capturing language information from text and then utilizing the information for specific NLP tasks. These pre-training approaches can be divided into two classes, i.e., feature-based approaches and fine-tuning approaches.

The early work~\cite{collobert2008unified,mikolov2013distributed,pennington2014glove} focuses on adopting feature-based approaches to transform words into distributed representations. As these pre-trained word representations capture syntactic and semantic information in textual corpora, they are often used as input embeddings and initialization parameters for various NLP models, and offer significant improvements over random initialization parameters~\cite{turian2010word}. Since these word-level models often suffer from the word polysemy, \newcite{peters2018deep} further adopt the sequence-level model (ELMo) to capture complex word features across different linguistic contexts and use ELMo to generate context-aware word embeddings.

Different from the above-mentioned feature-based language approaches only using the pre-trained language representations as input features, \newcite{dai2015semi} train auto-encoders on unlabeled text, and then use the pre-trained model architecture and parameters as a starting point for other specific NLP models. Inspired by~\newcite{dai2015semi}, more pre-trained language representation models for fine-tuning have been proposed. \newcite{howard2018universal} present AWD-LSTM~\cite{merity2017regularizing} to build a universal language model (ULMFiT).  \newcite{radford2018improving} propose a generative pre-trained Transformer~\cite{vaswani2017attention} (GPT) to learn language representations. \newcite{devlin2018bert} propose a deep bidirectional model with multiple-layer Transformers (BERT), which achieves the state-of-the-art results for various NLP tasks.

Though both feature-based and fine-tuning language representation models have achieved great success, they ignore the incorporation of knowledge information. As demonstrated in recent work, injecting extra knowledge information can significantly enhance original models, such as reading comprehension~\cite{mihaylov2018knowledgeable,zhong2018improving}, machine translation~\cite{zaremoodi2018adaptive}, natural language inference~\cite{chen2018neural}, knowledge acquisition~\cite{han2018neural}, and dialog systems~\cite{madotto2018mem2seq}. Hence, we argue that extra knowledge information can effectively benefit existing pre-training models. In fact, some work has attempted to joint representation learning of words and entities for effectively leveraging external KGs and achieved promising results~\cite{wang2014knowledge,toutanova2015representing,han2016joint,yamada2016joint,cao2017bridge,cao2018joint}. \newcite{sun2019ernie} propose the knowledge masking strategy for masked language model to enhance language representation by knowledge~\footnote{It is a coincidence that both \newcite{sun2019ernie} and we chose ERNIE as the model names, which follows the interesting naming habits like ELMo and BERT. \newcite{sun2019ernie} released their code on March 16th and submitted their paper to Arxiv on April 19th while we submitted our paper to ACL whose deadline is March 4th.}. In this paper, we further utilize both corpora and KGs to train an enhanced language representation model based on BERT.

\section{Methodology}

In this section, we present the overall framework of ERNIE and its detailed implementation, including the model architecture in Section~\ref{sec:architecture}, the novel pre-training task designed for encoding informative entities and fusing heterogeneous information in Section~\ref{sec:loss}, and the details of the fine-tuning procedure in Section~\ref{sec:fine-tuning}.

\subsection{Notations}

We denote a token sequence as $\{w_1, \ldots, w_{n}\}$~\footnote{In this paper, tokens are at the subword level.}, where $n$ is the length of the token sequence. Meanwhile, we denote the entity sequence aligning to the given tokens as $\{e_1, \ldots, e_{m}\}$, where $m$ is the length of the entity sequence. Note that $m$ is not equal to $n$ in most cases, as not every token can be aligned to an entity in KGs. Furthermore, we denote the whole vocabulary containing all tokens as $\mathcal{V}$, and the entity list containing all entities in KGs as $\mathcal{E}$. If a token $w \in \mathcal{V}$ has a corresponding entity $e \in \mathcal{E}$, their alignment is defined as $f(w) = e$. In this paper, we align an entity to the first token in its named entity phrase, as shown in Figure~\ref{fig:fusion}.

\subsection{Model Architecture}
\label{sec:architecture}

As shown in Figure~\ref{fig:fusion}, the whole model architecture of ERNIE consists of two stacked modules: (1) the underlying textual encoder ($\texttt{T-Encoder}$) responsible to capture basic lexical and syntactic information from the input tokens, and (2) the upper knowledgeable encoder ($\texttt{K-Encoder}$) responsible to integrate extra token-oriented knowledge information into textual information from the underlying layer, so that we can represent heterogeneous information of tokens and entities into a united feature space. Besides, we denote the number of $\texttt{T-Encoder}$ layers as $N$, and the number of $\texttt{K-Encoder}$ layers as $M$.

To be specific, given a token sequence $\{w_1, \ldots, w_n\}$ and its corresponding entity sequence $\{e_1, \ldots, e_{m}\}$, the textual encoder firstly sums the token embedding, segment embedding, positional embedding for each token to compute its input embedding, and then computes lexical and syntactic features $\{\bm{w}_1, \ldots, \bm{w}_n\}$ as follows,
\begin{equation}
\small
\{\bm{w}_1, \ldots, \bm{w}_n\} = \texttt{T-Encoder}(\{w_1, \ldots, w_n\}),
\end{equation}
where $\texttt{T-Encoder}(\cdot)$ is a multi-layer bidirectional Transformer encoder. As $\texttt{T-Encoder}(\cdot)$ is identical to its implementation in BERT and BERT is prevalent, we exclude a comprehensive description of this module and refer readers to~\newcite{devlin2018bert} and~\newcite{vaswani2017attention}.

After computing $\{\bm{w}_1, \ldots, \bm{w}_n\}$, ERNIE adopts a knowledgeable encoder $\texttt{K-Encoder}$ to inject the knowledge information into language representation. To be specific, we represent $\{e_1, \ldots, e_{m}\}$ with their entity embeddings $\{\bm{e}_1, \ldots, \bm{e}_{m}\}$, which are pre-trained by the effective knowledge embedding model TransE~\cite{bordes2013translating}. Then, both $\{\bm{w}_1, \ldots, \bm{w}_n\}$ and $\{\bm{e}_1, \ldots, \bm{e}_{m}\}$ are fed into $\texttt{K-Encoder}$ for fusing heterogeneous information and computing final output embeddings,
\begin{equation}
\small
\begin{aligned}
\{\bm{w}_1^{o}, & \ldots, \bm{w}_n^{o} \},  \{\bm{e}_1^{o}, \ldots, \bm{e}_n^{o}\} = \texttt{K-Encoder}( \\ 
& \{\bm{w}_1, \ldots, \bm{w}_n\}, \{\bm{e}_1, \ldots, \bm{e}_m\} ).
\end{aligned}
\end{equation}
$\{\bm{w}_1^{o},  \ldots, \bm{w}_n^{o} \}$ and $\{\bm{e}_1^{o}, \ldots, \bm{e}_n^{o}\}$ will be used as features for specific tasks. More details of the knowledgeable encoder $\texttt{K-Encoder}$ will be introduced in Section~\ref{sec:k-encoder}. 

\subsection{Knowledgeable Encoder}
\label{sec:k-encoder}

As shown in Figure~\ref{fig:fusion}, the knowledgeable encoder $\texttt{K-Encoder}$ consists of stacked aggregators, which are designed for encoding both tokens and entities as well as fusing their heterogeneous features. In the $i$-th aggregator, the input token embeddings $\{\bm{w}_1^{(i-1)},\ldots,\bm{w}_n^{(i-1)}\}$ and entity embeddings $\{\bm{e}_1^{(i-1)},\ldots,\bm{e}_m^{(i-1)}\}$ from the preceding aggregator are fed into two multi-head self-attentions (\texttt{MH-ATTs})~\cite{vaswani2017attention} respectively,
\begin{equation}
\small
\begin{aligned}
\{\bm{\tilde{w}}_1^{(i)},\ldots,\bm{\tilde{w}}_n^{(i)}\} &= \texttt{MH-ATT}(\{\bm{{w}}_1^{(i-1)},\ldots,\bm{{w}}_n^{(i-1)}\}),\\
\{\bm{\tilde{e}}_1^{(i)},\ldots,\bm{\tilde{e}}_m^{(i)}\} &= \texttt{MH-ATT}(\{\bm{{e}}_1^{(i-1)},\ldots,\bm{{e}}_m^{(i-1)}\}).
\end{aligned}
\end{equation}

Then, the $i$-th aggregator adopts an information fusion layer for the mutual integration of the token and entity sequence, and computes the output embedding for each token and entity. For a token $w_j$ and its aligned entity $e_k = f(w_j)$, the information fusion process is as follows,
\begin{equation}
	\small
\begin{aligned}
	\bm{h}_j &= \sigma (\bm{\tilde{W}}_{t}^{(i)} \bm{\tilde{w}}^{(i)}_j + \bm{\tilde{W}}_{e}^{(i)} \bm{\tilde{e}}^{(i)}_k + \bm{\tilde{b}}^{(i)}),\\
	\bm{w}^{(i)}_j &= \sigma (\bm{W}_{t}^{(i)} \bm{h}_j + \bm{b}^{(i)}_{t}), \\
	\bm{e}^{(i)}_k &= \sigma (\bm{W}_{e}^{(i)} \bm{h}_j + \bm{b}^{(i)}_{e}). \\
\end{aligned}
\end{equation}
where $\bm{h}_j$ is the inner hidden state integrating the information of both the token and the entity. $\sigma(\cdot)$ is the non-linear activation function, which usually is the GELU function~\cite{hendrycks2016gaussian}. For the tokens without corresponding entities, the information fusion layer computes the output embeddings without integration as follows,
\begin{equation}
	\small
\begin{aligned}
	\bm{h}_j &= \sigma (\bm{\tilde{W}}_{t}^{(i)} \bm{\tilde{w}}^{(i)}_j + \bm{\tilde{b}}^{(i)}),\\
	\bm{w}^{(i)}_j &= \sigma (\bm{W}_{t}^{(i)} \bm{h}_j + \bm{b}^{(i)}_{t}). \\
\end{aligned}
\end{equation}

For simplicity, the $i$-th aggregator operation is denoted as follows,
\begin{equation}
\small
\begin{aligned}
\{\bm{w}^{(i)}_1, & \ldots, \bm{w}^{(i)}_n\},\{\bm{e}^{(i)}_1, \ldots, \bm{e}^{(i)}_m\} = \texttt{Aggregator} ( \\
&\{\bm{w}^{(i-1)}_1, \ldots, \bm{w}^{(i-1)}_n\},\{\bm{e}^{(i-1)}_1, \ldots, \bm{e}^{(i-1)}_m\}). 
\end{aligned}
\end{equation}
The output embeddings of both tokens and entities computed by the top aggregator will be used as the final output embeddings of the knowledgeable encoder $\texttt{K-Encoder}$.

\begin{figure*}[t]
\centering
\includegraphics[width = 0.8\linewidth]{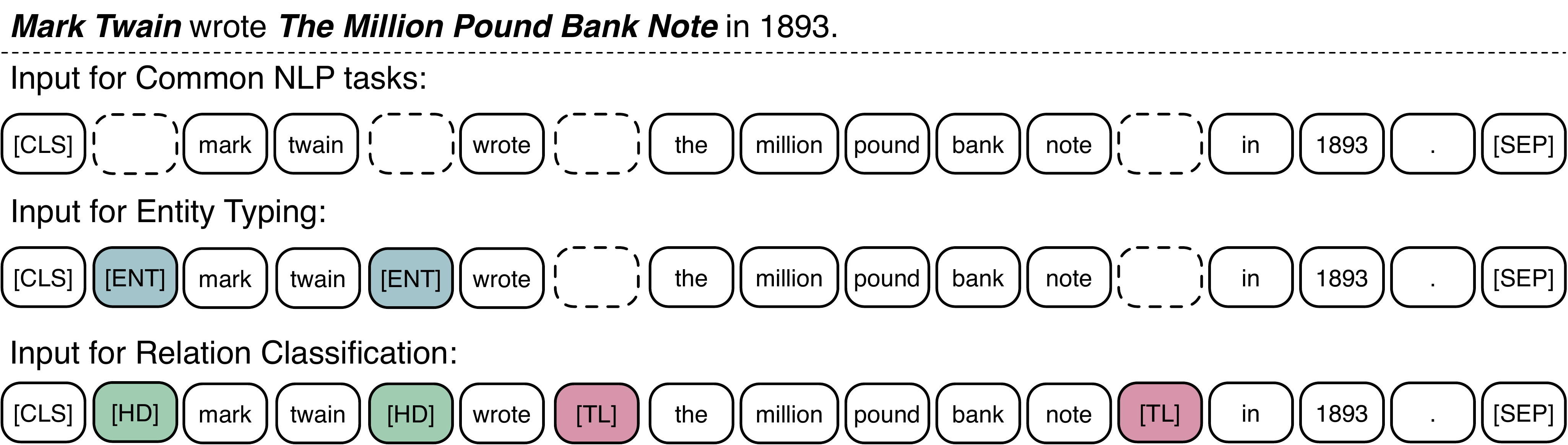}
\captionsetup{font={normalsize}}
\caption{Modifying the input sequence for the specific tasks. To align tokens among different types of input, we use dotted rectangles as placeholder. The colorful rectangles present the specific mark tokens.}
\label{fig:finetune}
\end{figure*}

\subsection{Pre-training for Injecting Knowledge}
\label{sec:loss}

In order to inject knowledge into language representation by informative entities, we propose a new pre-training task for ERNIE, which randomly masks some
token-entity alignments and then requires the system to predict all corresponding entities based on aligned tokens. As our task is similar to training a denoising auto-encoder~\cite{vincent2008extracting}, we refer to this procedure as a denoising entity auto-encoder (dEA). Considering that the size of $\mathcal{E}$ is quite large for the softmax layer, we thus only require the system to predict entities based on the given entity sequence instead of all entities in KGs. Given the token sequence $\{w_1, \ldots, w_n\}$ and its corresponding entity sequence $\{e_1, \ldots, e_{m}\}$, we define the aligned entity distribution for the token $w_i$ as follows,
\begin{equation}
\small
\label{eq:entity_prediction}
p(e_j|w_i) = \frac{\exp(\texttt{linear}(\bm{w}_i^o) \cdot \bm{e}_j)}{\sum_{k=1}^{m} \exp(\texttt{linear}(\bm{w}_i^o) \cdot \bm{e}_k) }, \\
\end{equation}
where $\texttt{linear}(\cdot)$ is a linear layer. Eq.~\ref{eq:entity_prediction} will be used to compute the cross-entropy loss function for dEA.

Considering that there are some errors in token-entity alignments, we perform the following operations for dEA: (1) In $5\%$ of the time, for a given token-entity alignment, we replace the entity with another random entity, which aims to train our model to correct the errors that the token is aligned with a wrong entity; (2) In $15\%$ of the time, we mask token-entity alignments, which aims to train our model to correct the errors that the entity alignment system does not extract all existing alignments; (3) In the rest of the time, we keep token-entity alignments unchanged, which aims to encourage our model to integrate the entity information into token representations for better language understanding.

Similar to BERT, ERNIE also adopts the masked language model (MLM) and the next sentence prediction (NSP) as pre-training tasks to enable ERNIE to capture lexical and syntactic information from tokens in text. More details of these pre-training tasks can be found from~\newcite{devlin2018bert}. The overall pre-training loss is the sum of the dEA, MLM and NSP loss.




\subsection{Fine-tuning for Specific Tasks}
\label{sec:fine-tuning}


As shown in Figure~\ref{fig:finetune}, for various common NLP tasks, ERNIE can adopt the fine-tuning procedure similar to BERT. We can take the final output embedding of the first token, which corresponds to the special [CLS] token, as the representation of the input sequence for specific tasks. For some knowledge-driven tasks (e.g., relation classification and entity typing), we design special fine-tuning procedure:

For relation classification, the task requires systems to classify relation labels of given entity pairs based on context. The most straightforward way to fine-tune ERNIE for relation classification is to apply the pooling layer to the final output embeddings of the given entity mentions, and represent the given entity pair with the concatenation of their mention embeddings for classification. In this paper, we design another method, which modifies the input token sequence by adding two mark tokens to highlight entity mentions. These extra mark tokens play a similar role like position embeddings in the conventional relation classification models~\cite{zeng2015distant}. Then, we also take the [CLS] token embedding for classification. Note that we design different tokens \textbf{[HD]} and \textbf{[TL]} for head entities and tail entities respectively.

The specific fine-tuning procedure for entity typing is a simplified version of relation classification. As previous typing models make full use of both context embeddings and entity mention embeddings~\cite{shimaoka2016attentive,yaghoobzadeh2017multi,xin2018put}, we argue that the modified input sequence with the mention mark token \textbf{[ENT]} can guide ERNIE to combine both context information and entity mention information attentively.

\section{Experiments}

In this section, we present the details of pre-training ERNIE and the fine-tuning results on five NLP datasets, which contain both knowledge-driven tasks and the common NLP tasks.

\subsection{Pre-training Dataset}

The pre-training procedure primarily acts in accordance with the existing literature on pre-training language models. For the large cost of training ERNIE from scratch, we adopt the parameters of BERT released by Google\footnote{https://github.com/google-research/bert} to initialize the Transformer blocks for encoding tokens. Since pre-training is a multi-task procedure consisting of NSP, MLM, and dEA, we use English Wikipedia as our pre-training corpus and align text to Wikidata. After converting the corpus into the formatted data for pre-training, the annotated input has nearly $4,500$M subwords and $140$M entities, and discards the sentences having less than $3$ entities.

Before pre-training ERNIE, we adopt the knowledge embeddings trained on Wikidata\footnote{https://www.wikidata.org/} by TransE as the input embeddings for entities. To be specific, we sample part of Wikidata which contains $5,040,986$ entities and $24,267,796$ fact triples. The entity embeddings are fixed during training and the parameters of the entity encoding modules are all initialized randomly.

\subsection{Parameter Settings and Training Details}

In this work, we denote the hidden dimension of token embeddings and entity embeddings as $H_w$, $H_e$ respectively, and the number of self-attention heads as $A_w$, $A_e$ respectively. In detail, we have the following model size: $N=6, M=6, H_w=768,H_e=100,A_w=12,A_e=4$. The total parameters are about $114$M.

The total amount of parameters of $\text{BERT}_{\text{BASE}}$ is about $110$M, which means the knowledgeable module of ERNIE is much smaller than the language module and has little impact on the run-time performance. And, we only pre-train ERNIE on the annotated corpus for one epoch. To accelerate the training process, we reduce the max sequence length from $512$ to $256$ as the computation of self-attention is a quadratic function of the length. To keep the number of tokens in a batch as same as BERT, we double the batch size to $512$. Except for setting the learning rate as $5e^{-5}$, we largely follow the pre-training hyper-parameters used in BERT.
For fine-tuning, most hyper-parameters are the same as pre-training, except batch size, learning rate, and number of training epochs. We find the following ranges of possible values work well on the training datasets with gold annotations, i.e., batch size: 32, learning rate (Adam): $5e^{-5}$, $3e^{-5}$, $2e^{-5}$, number of epochs ranging from 3 to 10.

\begin{table}[t]
\centering
\scriptsize
\begin{tabular}{l|rrrr}
  \toprule
  Dataset & Train & Develop & Test &Type\\
  \midrule
  FIGER & 2,000,000 & 10,000 & 563 & 113\\
  Open Entity & 2,000 & 2,000 & 2,000 & 6\\
  \bottomrule
\end{tabular}
\caption{The statistics of the entity typing datasets FIGER and Open Entity.}
\label{tab-typing}
\end{table}

\begin{table}[t]
\scriptsize
\centering
\begin{tabular}{l|rrr}
  \toprule
  Model & Acc.& Macro & Micro\\
  \midrule
  NFGEC (Attentive) & 54.53 & 74.76 & 71.58 \\
  NFGEC (LSTM) & 55.60 & 75.15 & 71.73 \\
  BERT & 52.04 & 75.16 & 71.63 \\
  \midrule
  ERNIE & \textbf{57.19} & \textbf{76.51} & \textbf{73.39} \\
  \bottomrule
\end{tabular}
\caption{Results of various models on FIGER (\%).}
\label{tab-et-result}
\vspace{-3mm}
\end{table}

We also evaluate ERNIE on the distantly supervised dataset, i.e., FIGER~\cite{ling2015design}. As the powerful expression ability of deeply stacked Transformer blocks, we found small batch size would lead the model to overfit the training data. Hence, we use a larger batch size and less training epochs to avoid overfitting, and keep the range of learning rate unchanged, i.e., batch size: 2048, number of epochs: 2, 3.

As most datasets do not have entity annotations, we use TAGME~\cite{ferragina2010tagme} to extract the entity mentions in the sentences and link them to their corresponding entities in KGs. 

\subsection{Entity Typing}

Given an entity mention and its context, entity typing requires systems to label the entity mention with its respective semantic types. To evaluate performance on this task, we fine-tune ERNIE on two well-established datasets FIGER~\cite{ling2015design} and Open Entity~\cite{choi2018ultra}. The training set of FIGER is labeled with distant supervision, and its test set is annotated by human. Open Entity is a completely manually-annotated dataset. The statistics of these two datasets are shown in Table~\ref{tab-typing}. We compare our model with the following baseline models for entity typing:

\begin{table}[t]
\centering
\scriptsize
\begin{tabular}{l|rrr}
  \toprule
  Model & P & R & F1\\
  \midrule
  NFGEC (LSTM) & 68.80 & 53.30 & 60.10\\
  UFET & 77.40 & 60.60 & 68.00\\
  BERT & 76.37 & 70.96 & 73.56 \\
  \midrule
  ERNIE & \textbf{78.42} & \textbf{72.90} & \textbf{75.56} \\
  \bottomrule
\end{tabular}
\caption{Results of various models on Open Entity (\%).}
\label{tab-et-open}
\vspace{-3mm}
\end{table}

\paragraph{NFGEC.} NFGEC is a hybrid model proposed by~\newcite{shimaoka2016attentive}. NFGEC combines the representations of entity mention, context and extra hand-craft features as input, and is the state-of-the-art model on FIGER. As this paper focuses on comparing the general language representation abilities of various neural models, we thus do not use the hand-craft features in this work. 

\paragraph{UFET.} For Open Entity, we add a new hybrid model UFET~\cite{choi2018ultra} for comparison. UFET is proposed with the Open Entity dataset, which uses a Bi-LSTM for context representation instead of two Bi-LSTMs separated by entity mentions in NFGEC. 

Besides NFGEC and UFET, we also report the result of fine-tuning BERT with the same input format introduced in Section~\ref{sec:fine-tuning} for fair comparison. Following the same evaluation criteria used in the previous work, we compare NFGEC, BERT, ERNIE on FIGER, and adopt strict accuracy, loose macro, loose micro scores for evaluation. We compare NFGEC, BERT, UFET, ERNIE on Open Entity, and adopt precision, recall, micro-F1 scores for evaluation.

The results on FIGER are shown in Table~\ref{tab-et-result}. From the results, we observe that: (1) BERT achieves comparable results with NFGEC on the macro and micro metrics. However, BERT has lower accuracy than the best NFGEC model. As strict accuracy is the ratio of instances whose predictions are identical to human annotations, it illustrates some wrong labels from distant supervision are learned by BERT due to its powerful fitting ability. (2) Compared with BERT, ERNIE significantly improves the strict accuracy, indicating the external knowledge regularizes ERNIE to avoid fitting the noisy labels and accordingly benefits entity typing.

The results on Open Entity are shown in Table~\ref{tab-et-open}. From the table, we observe that: (1) BERT and ERNIE achieve much higher recall scores than the previous entity typing models, which means pre-training language models make full use of both the unsupervised pre-training and manually-annotated training data for better entity typing. (2) Compared to BERT, ERNIE improves the precision by $2\%$ and the recall by $2\%$, which means the informative entities help ERNIE predict the labels more precisely.

\begin{table}[t]
\centering
\scriptsize
\begin{tabular}{lcccc}
  \toprule
  Dataset & Train & Develop & Test & Relation \\
  \midrule
  FewRel & 8,000 & 16,000  & 16,000 & 80 \\
  TACRED & 68,124 & 22,631 & 15,509 & 42 \\
  \bottomrule
\end{tabular}
\caption{The statistics of the relation classification datasets FewRel and TACRED.}
\label{tab-re}
\vspace{-3mm}
\end{table}

In summary, \textbf{ERNIE effectively reduces the noisy label challenge in FIGER, which is a widely-used distantly supervised entity typing dataset, by injecting the information from KGs.} Besides, ERNIE also outperforms the baselines on Open Entity which has gold annotations. 

\subsection{Relation Classification}

Relation classification aims to determine the correct relation between two entities in a given sentence, which is an important knowledge-driven NLP task. To evaluate performance on this task, we fine-tune ERNIE on two well-established datasets FewRel~\cite{han2018fewrel} and TACRED~\cite{zhang2017position}. The statistics of these two datasets are shown in Table~\ref{tab-re}. As the original experimental setting of FewRel is few-shot learning, we rearrange the FewRel dataset for the common relation classification setting. Specifically, we sample $100$ instances from each class for the training set, and sample $200$ instances for the development and test respectively. There are $80$ classes in FewRel, and there are $42$ classes (including a special relation \textit{``no relation''}) in TACRED. We compare our model with the following baseline models for relation classification:

\paragraph{CNN.} With a convolution layer, a max-pooling layer, and a non-linear activation layer, CNN gets the output sentence embedding, and then feeds it into a relation classifier. To better capture the position of head and tail entities, position embeddings are introduced into CNN~\cite{zeng2015distant,lin2016neural,wu2017adversarial,han2018hierarchical}.

\paragraph{PA-LSTM.} \newcite{zhang2017position} propose PA-LSTM introducing a position-aware attention mechanism over an LSTM network, which evaluates the relative contribution of each word in the sequence for the final sentence representation.

\paragraph{C-GCN.} \newcite{zhang2018graph} adopt the graph convolution operations to model dependency trees for relation classification. To encode the word order and reduce the side effect of errors in dependency parsing, Contextualized GCN (C-GCN) firstly uses Bi-LSTM to generate contextualized representations as input for GCN models.

\begin{table}[t]
\scriptsize
\centering
\begin{tabular}{l|rrr|rrr}
  \toprule
  \multirow{2}{*}{Model} & \multicolumn{3}{c|}{FewRel} & \multicolumn{3}{c}{TACRED} \\
  & P & R & F1 & P & R & F1\\
  \midrule
  CNN & 69.51 & 69.64 & 69.35 & 70.30 & 54.20 & 61.20\\ 
  PA-LSTM &- & - & - & 65.70 & 64.50 & 65.10 \\
  C-GCN &- &- &- & 69.90 & 63.30 & 66.40 \\
  BERT & 85.05 & 85.11 & 84.89& 67.23 & 64.81 & 66.00\\ 
  \midrule
  ERNIE & 88.49 & 88.44 & \textbf{88.32}& 69.97 & 66.08 & \textbf{67.97}\\
  \bottomrule
\end{tabular}
\caption{Results of various models on FewRel and TACRED (\%).}
\label{tab-re-res}
\vspace{-3mm}
\end{table}

In addition to these three baselines, we also fine-tune BERT with the same input format introduced in Section~\ref{sec:fine-tuning} for fair comparison.

As FewRel does not have any null instance where there is not any relation between entities, we adopt macro averaged metrics to present the model performances. Since FewRel is built by checking whether the sentences contain facts in Wikidata, we drop the related facts in KGs before pre-training for fair comparison. From Table~\ref{tab-re-res}, we have two observations: (1) As the training data does not have enough instances to train the CNN encoder from scratch, CNN just achieves an F1 score of $69.35\%$. However, the pre-training models including BERT and ERNIE increase the F1 score by at least $15\%$. (2) ERNIE achieves an absolute F1 increase of $3.4\%$ over BERT, which means fusing external knowledge is very effective.

\begin{table}[t]
\scriptsize
\centering
\begin{tabular}{l|cccc}
  \toprule
  Model & MNLI-(m/mm) & QQP & QNLI & SST-2 \\
   & 392k & 363k & 104k & 67k\\
  \midrule
  $\text{BERT}_{\text{BASE}}$ & 84.6/83.4 & 71.2 & - & 93.5 \\
  \midrule
  ERNIE & 84.0/83.2 & 71.2 & 91.3 & 93.5 \\
  \midrule
  \midrule
  Model & CoLA & STS-B & MRPC & RTE \\
   &  8.5k & 5.7k & 3.5k & 2.5k \\
  \midrule
  $\text{BERT}_{\text{BASE}}$ & 52.1 & 85.8 & 88.9 & 66.4 \\
  \midrule
  ERNIE & 52.3 & 83.2 & 88.2 & 68.8 \\
  \bottomrule
\end{tabular}
\caption{Results of BERT and ERNIE on different tasks of GLUE (\%).}
\label{tab-glue}
\vspace{-3mm}
\end{table}

In TACRED, there are nearly $80\%$ null instances so that we follow the previous work~\cite{zhang2017position} to adopt micro averaged metrics to represent the model performances instead of the macro.
 The results of CNN, PA-LSTM, and C-GCN come from the paper by~\newcite{zhang2018graph}, which are the best results of CNN, RNN, and GCN respectively. From Table~\ref{tab-re-res}, we observe that: (1) The C-GCN model outperforms the strong BERT model by an F1 increase of $0.4\%$, as C-GCN utilizes the dependency trees and the entity mask strategy. The entity mask strategy refers to replacing each subject (and object similarly) entity with a special NER token, which is similar to our proposed pre-training task dEA. (2) ERNIE achieves the best recall and F1 scores, and increases the F1 of BERT by nearly $2.0\%$, which proves the effectiveness of the knowledgeable module for relation classification. 


In conclusion, we find that the pre-trained language models can provide more information for relation classification than the vanilla encoder CNN and RNN. And ERNIE outperforms BERT on both of the relation classification datasets, especially on the FewRel which has a much smaller training set. \textbf{It demonstrates extra knowledge helps the model make full use of small training data, which is important for most NLP tasks as large-scale annotated data is unavailable.}


\subsection{GLUE}

The General Language Understanding Evaluation (GLUE) benchmark~\cite{wang2018glue} is a collection of diverse natural language understanding tasks~\cite{warstadt2018neural, socher2013recursive, dolan2005automatically, agirre2007semantic, williams2018broad, rajpurkar2016squad, dagan2006pascal, levesque2011winograd}, which is the main benchmark used in~\newcite{devlin2018bert}. To explore whether our knowledgeable module degenerates the performance on common NLP tasks, we evaluate ERNIE on $8$ datasets of GLUE and compare it with BERT.

In Table~\ref{tab-glue}, we report the results of our evaluation submissions and those of BERT from the leaderboard. We notice that ERNIE is consistent with $\text{BERT}_{\text{BASE}}$ on big datasets like MNLI, QQP, QNLI, and SST-2. The results become more unstable on small datasets, that is, ERNIE is better on CoLA and RTE, but worse on STS-B and MRPC.

In short, ERNIE achieves comparable results with $\text{BERT}_{\text{BASE}}$ on GLUE. On the one hand, it means GLUE does not require external knowledge for language representation. On the other hand, it illustrates ERNIE does not lose the textual information after heterogeneous information fusion.

\subsection{Ablation Study}

\begin{table}[t]
\centering
\scriptsize
\begin{tabular}{l|rrr}
  \toprule
  Model & P & R & F1\\
  \midrule
  BERT & 85.05 & 85.11 & 84.89 \\
  \midrule
  ERNIE & 88.49 & 88.44 & \textbf{88.32} \\
  \ \ \ \ w/o entities & 85.89 & 85.89 & 85.79 \\
  \ \ \ \ w/o dEA & 85.85 & 85.75 & 85.62 \\
  \bottomrule
\end{tabular}
\caption{Ablation study on FewRel (\%).}
\label{tab-re-ablation}
\vspace{-3mm}
\end{table}

In this subsection, we explore the effects of the informative entities and the knowledgeable pre-training task (dEA) for ERNIE using FewRel dataset. \textbf{w/o entities} and \textbf{w/o dEA} refer to fine-tuning ERNIE without entity sequence input and the pre-training task dEA respectively. As shown in Table \ref{tab-re-ablation}, we have the following observations: (1) Without entity sequence input, dEA still injects knowledge information into language representation during pre-training, which increases the F1 score of BERT by $0.9\%$. (2) Although the informative entities bring much knowledge information which intuitively benefits relation classification, ERNIE without dEA takes little advantage of this, leading to the F1 increase of $0.7\%$.

\section{Conclusion}

In this paper, we propose ERNIE to incorporate knowledge information into language representation models. Accordingly, we propose the knowledgeable aggregator and the pre-training task dEA for better fusion of heterogeneous information from both text and KGs. The experimental results demonstrate that ERNIE has better abilities of both denoising distantly supervised data and fine-tuning on limited data than BERT. There are three important directions remain for future research: (1) inject knowledge into feature-based pre-training models such as ELMo~\cite{peters2018deep}; (2) introduce diverse structured knowledge into language representation models such as ConceptNet~\cite{speer2012representing} which is different from the world knowledge database Wikidata; (3) annotate more real-world corpora heuristically for building larger pre-training data. These directions may lead to more general and effective language understanding.


\section*{Acknowledgement}

This work is funded by the Natural Science Foundation of China (NSFC) and the German Research Foundation (DFG) in Project Crossmodal Learning, NSFC 61621136008 / DFG TRR-169, the National Natural Science Foundation of China (NSFC No. 61572273) and China Association for Science and Technology (2016QNRC001).

\small
\bibliography{acl2019}
\bibliographystyle{acl_natbib}

\end{document}